\documentclass[preprint,12pt,3p]{elsarticle}

\usepackage{amssymb}

\usepackage{graphicx}
\usepackage{amsmath,amssymb} 
\usepackage{xcolor}
\usepackage{color}
\usepackage{booktabs} 
\usepackage{times}
\usepackage{epsfig}
\usepackage{subfigure}
\usepackage{siunitx}
\usepackage{textcomp}
\usepackage[utf8]{inputenc}
\usepackage[english]{babel}
\usepackage{url}
\usepackage{colortbl}
\usepackage[caption=false]{subfig}
\usepackage{caption}

\DeclareMathOperator*{\argmin}{arg\,min}
\renewcommand{\refname}{\normalfont\selectfont\normalsize Publications}

\journal{Multi-Modal Scene Understanding}

\begin{document}
\begin{frontmatter}
\title{Self-Supervised Learning from Web Data for Multimodal Retrieval}

\author[label1,label2]{Raul Gomez\corref{cor1}}
\address[label1]{Eurecat, Centre Tecnològic de Catalunya, Unitat de Tecnologies Audiovisuals, Barcelona, Spain}
\address[label2]{Computer Vision Center, Universitat Autònoma de Barcelona, Barcelona, Spain}
\cortext[cor1]{Corresponding author}
\ead{raul.gomez@cvc.uab.es}

\author[label2]{Lluis Gomez}
\ead{lgomez@cvc.uab.es}

\author[label1]{Jaume Gibert}
\ead{jaume.gibert@eurecat.org}

\author[label2]{Dimosthenis Karatzas}
\ead{dimos@cvc.uab.es}

\begin{abstract}
Self-Supervised learning from multimodal image and text data allows deep neural networks to learn powerful features with no need of human annotated data. Web and Social Media platforms provide a virtually unlimited amount of this multimodal data.
In this work we propose to exploit this free available data to learn a multimodal image and text embedding, aiming to leverage the semantic knowledge learnt in the text domain and transfer it to a visual model for semantic image retrieval. 
We demonstrate that the proposed pipeline can learn from images with associated text without supervision and analyze the semantic structure of the learnt joint image and text embedding space.
We perform a thorough analysis and performance comparison of five different state of the art text embeddings in three different benchmarks. We show that the embeddings learnt with Web and Social Media data have competitive performances over supervised methods in the text based image retrieval task, and we clearly outperform state of the art in the MIRFlickr dataset when training in the target data.
Further, we demonstrate how semantic multimodal image retrieval can be performed using the learnt embeddings, going beyond classical instance-level retrieval problems.
Finally, we present a new dataset, InstaCities1M, composed by Instagram images and their associated texts that can be used for fair comparison of image-text embeddings.
\end{abstract}


\end{frontmatter}



\section{Introduction}

\subsection{Annotating Data: A Bottleneck for Training Deep Neural Networks}
Large annotated datasets, powerful hardware and deep learning techniques are allowing to get outstanding machine learning results. Not only in traditional classification problems but also in more challenging tasks such as image captioning or language translation. Deep neural networks allow building pipelines that can learn patterns from any kind of data with impressive results.

Deep learning has two strong requirements: Computation power and tons of data. The computation power requirement is fulfilled by GPUs and other AI specialized hardware, such as TPUs. Moreover, the hardware power is evolving fast without an apparent roof together with deep learning algorithms requirements. The story with the data requirement is different. 
Despite the existence of large-scale annotated datasets such as ImageNet \cite{Deng}, COCO \cite{Lin2014} or Places \cite{Zhou2017}, the lack of data limits the application of deep learning to specific problems where it is difficult or economically non-viable to get proper annotations.
Although there exist some tools to facilitate human data annotation, such as Amazon Mechanical Turk\footnote{\url{https://www.mturk.com}}, annotating the tons of data required to train supervised deep learning models is a very expensive and manual task, whose efficiency cannot evolve over time.

\subsection{Alternatives to Annotated Data}
A common strategy to overcome the lack of annotated data is to first train models in generic datasets, as ImageNet, and then fine-tune them to other tasks using smaller, specific datasets \cite{Yosinski}. But still we depend on the existence of annotated data to train our models. 
Another strategy to overcome the insufficiency of data is to use computer graphics techniques to generate artificial data inexpensively. However, while synthetic data has proven to be a valuable source of training data for many applications such as pedestrian detection \cite{Marin}, 
image semantic segmentation \cite{Ros2016} 
and scene text detection and recognition \cite{Jaderberg2014,Gupta2016}, nowadays it is still not easy to generate realistic complex images for some tasks. 

An alternative to this strategies and a solution to overcome the annotated data requirements of supervised deep learning techniques are not fully supervised techniques. Among them, self-supervised learning exploits multimodal data to learn relations between two or more data modalities using paired instances. 
Web and Social Media offer an immense amount of images accompanied with other information such as the image title, description or date. This data is noisy and unstructured but it is free and nearly unlimited. 
We mentioned that data annotation efficiency does not improve with time. As a contrast, the amount of available multi-modal data in the Web does. Designing algorithms to learn from Web data is an interesting research area as it would disconnect the deep learning evolution from the scaling of human-annotated datasets, given the enormous amount of existing Web and Social Media data. We call this scenario self-supervised learning because it consists in exploiting relations between different modalities (in this case images and text) of multimodal data as supervision.

\subsection{Exploiting Multimodal Web Data}

\begin{figure}[ht]
  \includegraphics[width=1\linewidth]{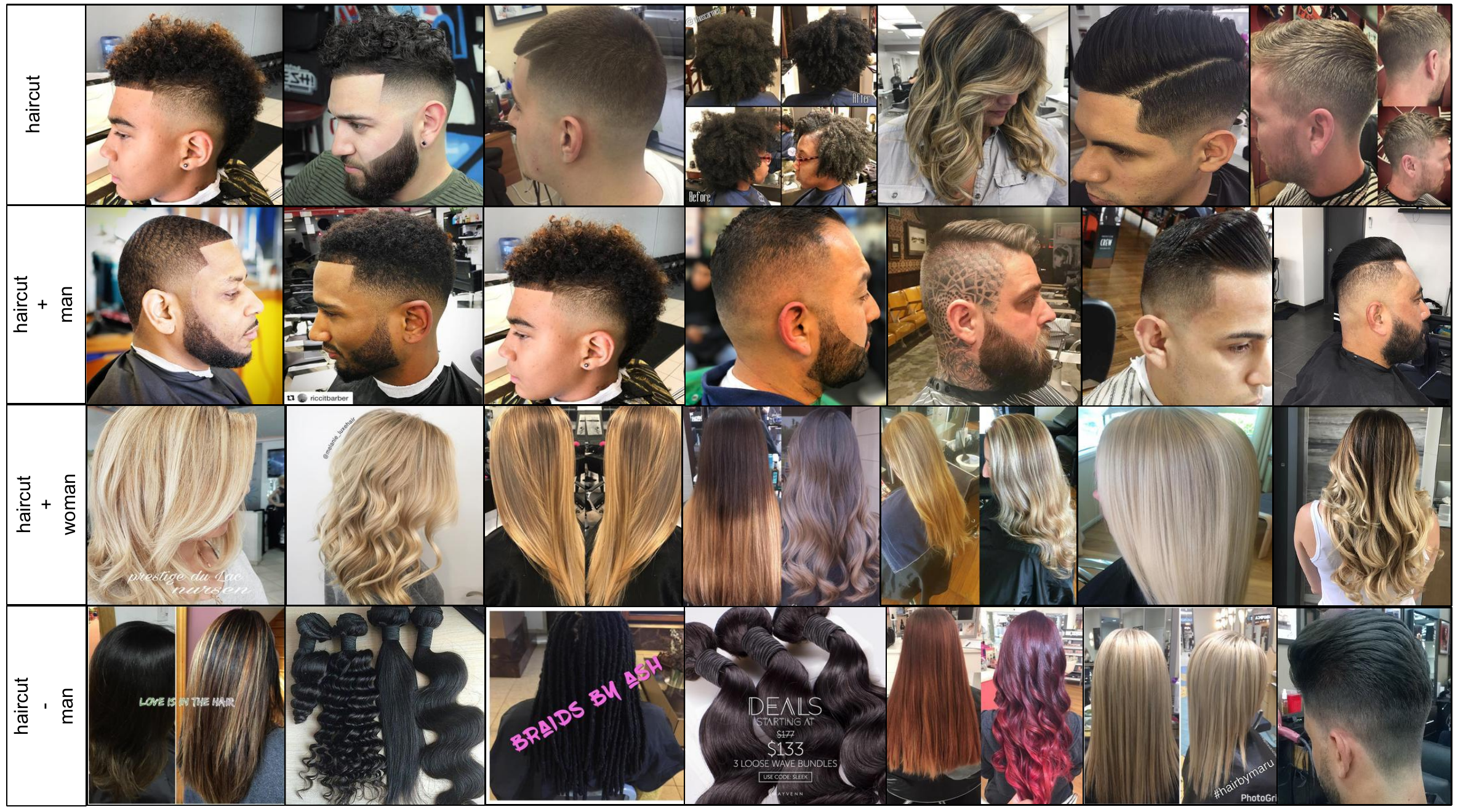}
   \caption{Top-ranked results of combined text queries by our semantic image retrieval model. The learnt joint image-text embedding permits to learn a rich semantic manifold even for previously unseen concepts even though they might not be explicitly present in the training set.}
   \label{fig:haircut}
\end{figure}

Lately, Web data has been used to build classification datasets, such as in the WebVision Challenge \cite{Li2017a} and in this Facebook work \cite{Mahajan}. In these works, to build a classification dataset, queries are made to search engines using class names and the retrieved images are labeled with the querying class. In such a configuration the learning is limited to some pre-established classes, thus it could not generalize to new classes. While working with image labels is very convenient for training traditional visual models, the semantics in such a discrete space are very limited in comparison with the richness of human language expressiveness when describing an image. 
Instead we define here a scenario where, by exploiting distributional semantics in a given text corpus, we can learn from every word associated to an image. As illustrated in Figure~\ref{fig:haircut}, by leveraging the richer semantics encoded in the learnt embedding space, we can infer previously unseen concepts even though they might not be explicitly present in the training set. 

The noisy and unstructured text associated to Web images provides information about the image content that we can use to learn visual features. A strategy to do that is to embed the multimodal data (images and text) in the same vectorial space. 
In this work we represent text using five different state of the art methods and eventually embed images in the learnt semantic space by means of a regression CNN. We compare the performance of the different text space configurations under a text based image retrieval task.

\section{Related Work}
Multimodal image and text embeddings have been lately a very active research area. The possibilities of learning together from different kinds of data have motivated this field of study, where both general and applied research has been done. 
DeViSE \cite{Frome2013} proposes a pipeline that, instead of learning to predict ImageNet classes, learns to infer the Word2Vec \cite{Mikolov2013} representations of their labels. The result is a model that makes semantically relevant predictions even when it makes errors, and generalizes to classes outside of its labeled training set.
Gordo \& Larlus \cite{DianeLarlus2017} use captions associated to images to learn a common embedding space for images and text through which they perform semantic image retrieval. They use a \textit{tf-idf} based BoW representation over the image captions as a semantic similarity measure between images and they train a CNN to minimize a margin loss based on the distances of triplets of query-similar-dissimilar images. Gomez, Patel \textit{et al.} \cite{Gomez2017, Patel2018} use LDA \cite{Blei2003} to extract topic probabilities from a bunch of Wikipedia articles and train a CNN to embed their associated images in the same topic space. Wang \textit{et al.} \cite{Wanga} propose a method to learn a joint embedding of images and text for image-to-text and text-to-image retrieval, by training a neural net to embed in the same space Word2Vec \cite{Mikolov2013} text representations and CNN extracted features. 

Other than semantic retrieval, joint image-text embeddings have also been used in more specific applications. Patel \textit{et al.} \cite{Patel} use LDA \cite{Blei2003} to learn a joint image-text embedding and generate contextualized lexicons for images using only visual information. Gordo \textit{et al.} \cite{Gordo} embed word images in a semantic space relying in the graph taxonomy provided by WordNet \cite{PrincetonUniversity2010} to perform text recognition. In a more specific application, Salvador \textit{et al.} \cite{Salvador} propose a joint embedding of food images and their recipes to identify ingredients, using Word2Vec \cite{Mikolov2013} and LSTM representations to encode ingredient names and cooking instructions and a CNN to extract visual features from the associated images. Exploiting Instagram publications related to \#Barcelona, Gomez \textit{et al.} \cite{Gomez2018a} learn relations between words, images and Barcelona neighbourhoods to study which words and visual features tourist and locals relate with each neighbourhood. 

The robustness against noisy data has also been addressed by the community, though usually in an implicit way. Patrini \textit{et al.} \cite{Patrini} address the problem of training a deep neural network with label noise with a loss correction approach and Xiau \textit{et al.} \cite{Xiao} propose a method to train a network with a limited number of clean labels and millions of noisy labels. Fu \textit{et al.} \cite{Fu} propose an image tagging method robust to noisy training data and Xu \textit{et al.} \cite{Xu2017SM} address social image tagging correction and completion. Zhang \textit{et al.} \cite{Zhang2016} show how label noise affects the CNN training process and its generalization error.

\subsection{Contributions}
The work presented here brings in a performance comparison between five state of the art text embeddings in self-supervised learning, showing results in three different datasets. Furthermore it proves that self-supervised multimodal learning can be applied to Web and Social Media data achieving competitive results in text-based image retrieval compared to pipelines trained with human annotated data. Finally, a new dataset formed by Instagram images and their associated text is presented: InstaCities1M.

\section{Multimodal Text-Image Embedding}

One of the objectives of this work is to serve as a fair comparative of different text embeddings methods when learning from Web and Social Media data. Therefore we design a pipeline to test the different methods under the same conditions, where the text embedding is a module that can be replaced by any text representation.

\begin{figure}[ht]
\centering
{\includegraphics[width=0.65\linewidth]{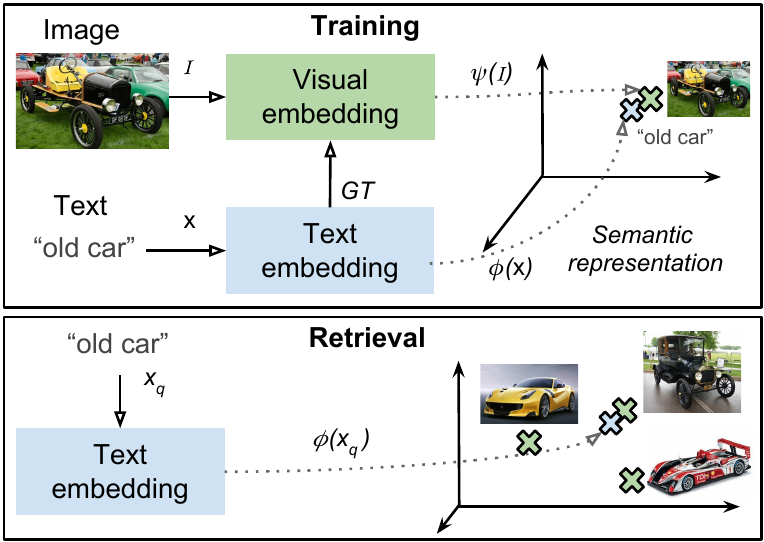}}
\caption{Pipeline of the visual embedding model training and the image retrieval by text.
\label{fig:pipeline}}
\end{figure}

The proposed pipeline is as follows: First, we train the text embedding model on a dataset composed by pairs of images and correlated texts $(I, x)$. Second, we use the text embedding model to generate vectorial representations of those texts. Given a text instance $x$, we denote its embedding by $\phi(x) \in \mathbb{R}^D$. Third, we train a CNN to regress those text embeddings directly from the correlated images. Given an image $I$, its representation in the embedding space is denoted by $\psi(I) \in \mathbb{R}^D$. Thereby the CNN learns to embed images in the vectorial space defined by the text embedding model.
The trained CNN model is used to generate visual embeddings for the test set images. Figure \ref{fig:pipeline} shows a diagram of the visual embedding training pipeline and the retrieval procedure.

In the image retrieval stage the vectorial representation in the joint text-image space of the querying text is computed using the text embedding model. Image queries can also be handled by using the visual embedding model instead of the text embedding model to generate the query representation. Furthermore, we can generate complex queries combining different query representations applying algebra in the joint text-image space. To retrieve the most semantically similar image $I_R$ to a query $x_q$, we compute the cosine similarity of its vectorial representation $\phi(x_q)$ with the visual embeddings of the test set images $\psi(I_T)$, and retrieve the nearest image in the joint text-image space:
\begin{equation}
\argmin_{I_T \in \text{Test}} \frac{ \langle \phi(x_q) , \psi(I_T) \rangle}{||\phi(x_q)|| \cdot ||\psi(I_T)||}.
\end{equation}
\vspace{5pt}

State of the art text embedding methods trained on large text corpus are very good generating representations of text in a vector space where semantically similar concepts fall close to each other. The proposed pipeline leverages the semantic structure of these text embedding spaces training a visual embedding model that generates vectorial representations of images in the same space, mapping semantically similar images close to each other, and also close to texts correlated to the image content.
Note that the proposed joint text-image embedding can be extended to other tasks besides image retrieval, such as image annotation, tagging or captioning.

A CNN is trained to regress text embeddings from the correlated images minimizing a sigmoid cross-entropy loss. This loss is used to minimize distances between the text and image embeddings. Let $\{(I_n, x_n)\}_{n=1:N}$ be a batch of image-text pairs.  If $\sigma(\cdot)$ is the component-wise sigmoid function, we denote $p_n = \sigma(\phi(x_n))$ and $\hat{p}_n = \sigma(\psi(I_n))$. Note ${p}_n$, $\hat{p}_n \in \mathbb{R}^D$ where $D$ is the dimensionality of the joint embedding space. Let the loss be:

\begin{equation}
L = -\tfrac{1}{ND}\sum_{n=1}^{N} \sum_{d=1}^{D} [\, p_{n_d} \log \hat{p}_{n_d} + (1-p_{n_d}) \log(1-\hat{p}_{n_d}) \, ],
\end{equation}
\vspace{5pt}

The GoogleNet architecture \cite{Szegedy} is used, customizing the last layer to regress a vector of  the same dimensionality as the text embedding. 
We train with a Stochastic Gradient Descent optimizer with a learning rate of $1e^{-3}$, multiplied by 0.1 every 100k iterations, and a momentum of 0.9. The batch size is set to 120 and random cropping and mirroring are used as online data augmentation. With these settings the CNN trainings converge after around 300K-500K iterations.
We use the Caffe \cite{Jia} framework and initialize with the ImageNet \cite{Deng} trained model to make the training faster. Notice that, despite initializing with a model trained with human-annotated data, this does not denote a dependence on annotated data, since the resulting model can generalize to much more concepts than the ImageNet classes. We trained one model from scratch obtaining similar results, although more training iterations were needed.
Cross Entropy Loss is not usually used for regression problems, where Mean Square Error loss is often used. We chose Cross Entropy Loss empirically, since it was the one providing an stable training and better performance. 
Although Cross Entropy Loss tends to be considered a loss for classification, it is also suitable for regression problems: despite this loss will not be zero when the regression solution matches the groundtruth, it will always be minimum compared to other solutions. 

\section{Text Embeddings}
Text vectorization methods are diverse in terms of architecture and the text structure they are designed to deal with. Some methods are oriented to vectorize individual words and others to vectorize full texts or paragraphs. In this work we consider the top-performing text embeddings and test them in our pipeline to evaluate their performance when learning from Web and Social Media data. Here we explain briefly the main characteristics of each text embedding method used.

\paragraph{LDA \cite{Blei2003}} Latent Dirichlet Allocation learns latent topics from a collection of text documents and maps words to a vector of probabilities of those topics. It can describe a document by assigning topic distributions to it, which in turn have word distributions assigned. An advantage of this method is that it gives interpretable topics.

\paragraph{GloVe \cite{Pennington}} It is a count-based model. It learns the vectors by essentially doing dimensionality reduction on the co-occurrence counts matrix. Training is performed on aggregated global word-word co-occurrence statistics from a corpus.

\paragraph{Word2Vec \cite{Mikolov2013}} Learns representations for words based on their context using a single hidden layer feed-forward neural network. 
It has two variants: In the CBOW (Continuous Bag of Word) approach, the neural network is trained to predict a word given as input its surrounding context (surrounding words). In the Skip-gram model, opposite to the CBOW model, the neural network is trained to predict a word context given that word as an input.  
In this work we use the most extended and efficient CBOW approach.

\paragraph{Doc2Vec \cite{Le2014}} Extends the Word2Vec idea to documents, being able to create a numeric representation for them, regardless of their length. Extending Word2Vec CBOW model, it adds another input vector to the input context, which is the paragraph identifier. When training the word vectors, the document vector is trained as well, and at the end it holds a numeric representation of the whole document.
As with Word2Vec, in this work we use the CBOW approach.

\paragraph{FastText \cite{Bojanowski2016}}  It is an extension of Word2Vec which treats each word as composed of character ngrams, learning representations for ngrams instead of words. The idea is to take into account and exploit the morphology of words. Each word is split in n-grams which are all inputted separately to the model, which can be trained using the CBOW or the skip-gram approach. The vector for each word is made of the sum of its character n grams, so it can generate embeddings for out of vocabulary words.
By exploiting words morphology, FastText tries to generate better embeddings for rare words, assuming their character ngrams are shared with other words. It also allows to generate embeddings for out of vocabulary words.
To train FastText we use the originaly proposed and most extended skigram approach. \\

To the best of our knowledge, this is the first time these text embeddings are trained from scratch on the same corpus and evaluated under the image retrieval by text task.
We used Gensim\footnote{\url{https://radimrehurek.com/gensim}} implementations of LDA, Word2Vec, FastText and Doc2Vec and the GloVe implementation by Maciej Kula\footnote{\url{https://github.com/maciejkula/glove-python}}.
While LDA and Doc2Vec can generate embeddings for documents, Word2Vec, GloVe and FastText only generate word embeddings. To get documents embeddings from these methods, we consider two standard strategies: First, computing the document embedding as the mean embedding of its words. Second, computing a \textit{tf-idf} weighted mean of the words in the document.
For all embeddings a dimensionality of 400 has been used. The value has been selected because is the one used in the Doc2Vec paper \cite{Le2014}, which compares Doc2Vec with other text embedding methods, and it is enough to get optimum performances of Word2Vec, FastText and GloVe, as \cite{Mikolov2013,Bojanowski2016,Pennington} show respectively.
For LDA a dimensionality of 200 has also been considered. 

\section{Benchmarks}
In this section we present the datasets used in this work and show some examples of their images and their associated text.

\subsection{InstaCities1M}
\begin{figure}[h]
\begin{center}
  \includegraphics[width=0.75\linewidth]{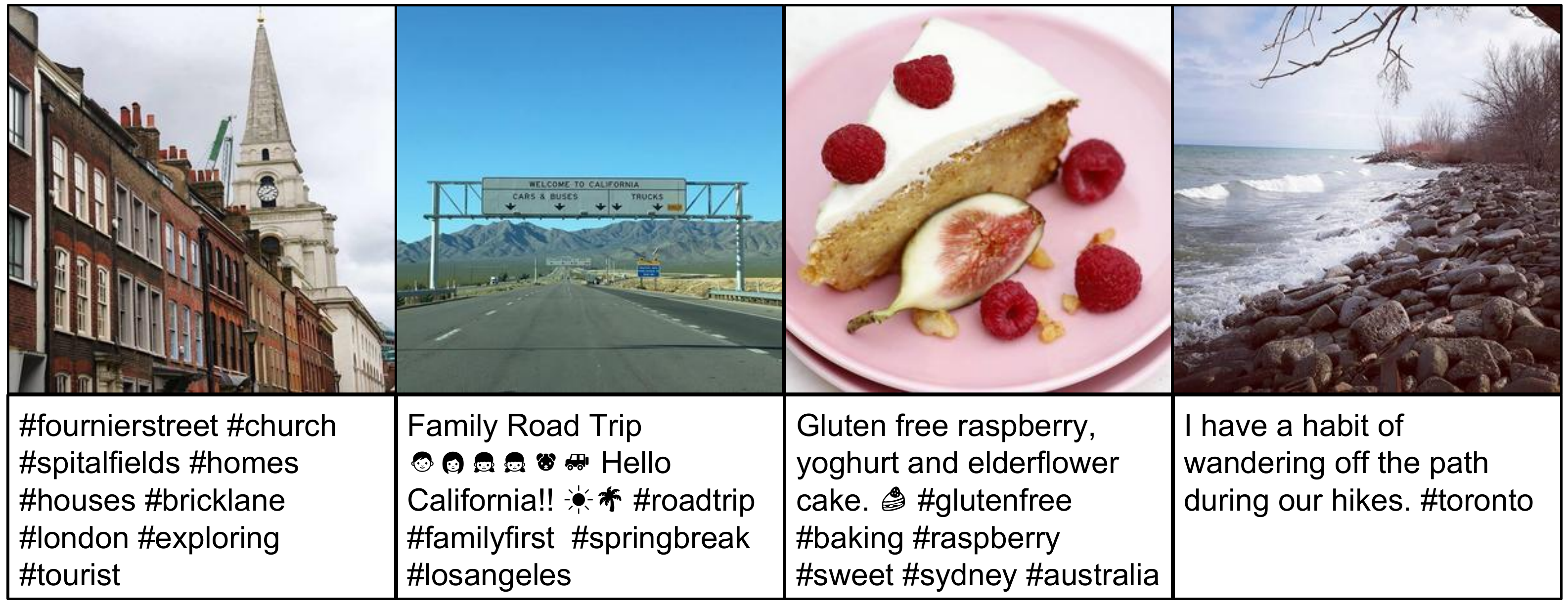}
   \caption{Examples of InstaCities1M dataset images.}
   \label{fig:exampes_InstaCities1M}
\end{center}
\end{figure}

A dataset formed by Instagram images associated with one of the 10 most populated English speaking cities all over the world (in the images captions one of the names of these cities appears). It contains 100K images for each city, which makes a total of 1M images, split in 800K training images, 50K validation images and 150K test images. The interest of this dataset is that is formed by recent Social Media data. The text associated with the images is the description and the hashtags written by the photo up-loaders, so it is the kind of free available data that would be very interesting to be able to learn from. Figure \ref{fig:exampes_InstaCities1M} shows some examples of InstaCities1M images and their associated text. The InstaCities1M dataset is available on \url{https://gombru.github.io/2018/08/01/InstaCities1M/}.

\subsection{WebVision}
\begin{figure}[h]
\begin{center}
  \includegraphics[width=0.75\linewidth]{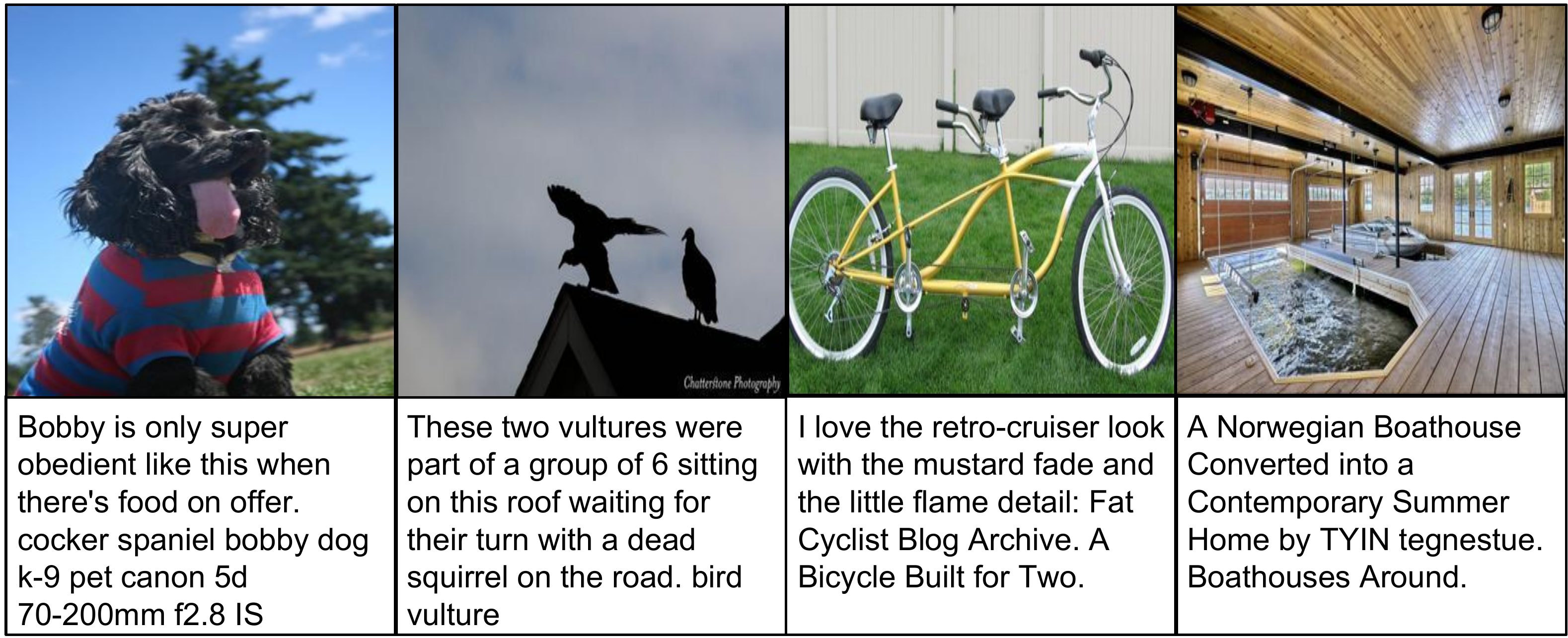}
   \caption{Examples of WebVision dataset images.}
   \label{fig:exampes_WebVision}
\end{center}
\end{figure}

The Webvision dataset \cite{Li2017} contains more than 2.4 million images crawled from the Flickr Website and Google Images search. The same 1,000 concepts as the ILSVRC 2012 dataset \cite{Deng} are used for querying images. The textual information accompanying those images (caption, user tags and description) is provided. The validation set, which is used as test in this work, contains 50K images. Figure \ref{fig:exampes_WebVision} shows some examples of WebVision images and their associated text.

\subsection{MIRFlickr}
\begin{figure}[h]
\begin{center}
  \includegraphics[width=0.75\linewidth]{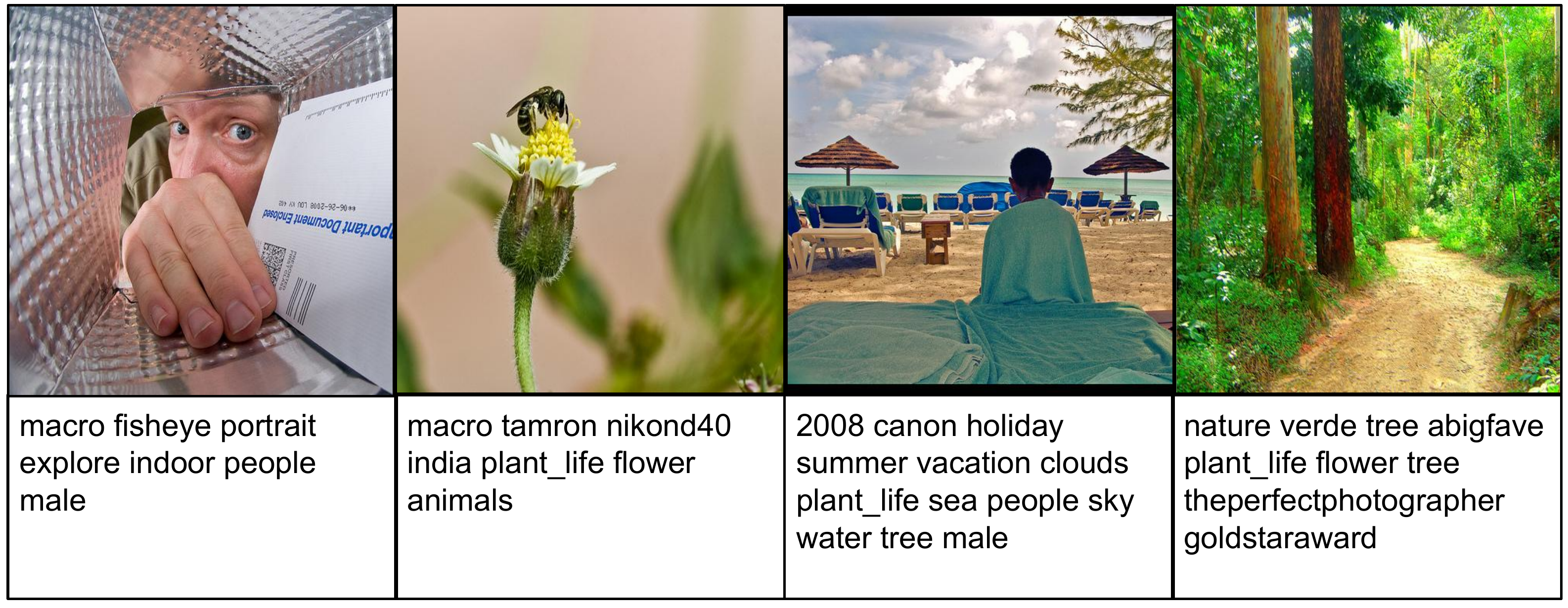}
   \caption{Examples of MIRFlickr dataset images.}
   \label{fig:exampes_MIRFlickr}
\end{center}
\end{figure}

The MIRFlickr dataset \cite{Huiskes} contains 25,000 images collected from Flickr, annotated using 24 predefined semantic concepts. 14 of those concepts are divided in two categories: 1) strong correlation concepts and 2) weak correlation concepts. The correlation between an image and a concept is strong if the concept appears in the image predominantly. For differentiation, we denote strong correlation concepts by a suffix ``*''. Finally, considering strong and weak concepts separately, we get 38 concepts in total. All images in the dataset are annotated by at least one of those concepts. Additionally, all images have associated tags collected from Flickr. 
Following the experimental protocol in \cite{Lin2015,Xu2017,Li2016,Liu2017} tags that appear less than 20 times are first removed and then instances without tags or annotations are removed. 
Figure \ref{fig:exampes_MIRFlickr} shows some examples of MIRFlickr images and their associated text.

\section{Retrieval on InstaCities1M and WebVision Datasets}
In this section we perform image retrieval experiments in the InstaCites1M and the WebVision datasets, comparing the performance of the different text embeddings in our pipeline. We analyze the performance of each text embedding, present an error analysis of our pipeline and show qualitative retrieval results of both image by text retrieval and image retrieval using multimodal queries.

\subsection{Experiment Setup}
To evaluate the learnt joint embeddings, we define a set of textual queries and check visually if the TOP-5 retrieved images contain the querying concept. 
We define 24 different queries. Half of them are single word queries and the other half two word queries. They have been selected to cover a wide area of semantic concepts that are usually present in Web and Social Media data. Both simple and complex queries are divided in four different categories: Urban, weather, food and people. 
Queries are listed in Table \ref{tab:queries}. For complex queries, only images containing both querying concepts are considered correct.\\

\begin{table}[h]
\begin{center}
\caption{Queries for the retrieval experiments on InstaCities1M and WebVision datasets.}
\resizebox{0.7\linewidth}{!}{
\begin{tabular}{l|l|l|}
\cline{2-3}
 & \multicolumn{1}{c|}{\textbf{Simple}} & \multicolumn{1}{c|}{\textbf{Complex}} \\ \hline
\multicolumn{1}{|l|}{\textbf{Urban}} & car, skyline, bike & yellow+car, skyline+night, bike+park \\ \hline
\multicolumn{1}{|l|}{\textbf{Weather}} & sunrise, snow, rain & sunrise+beach, snow+ski, rain+umbrella \\ \hline
\multicolumn{1}{|l|}{\textbf{Food}} & ice-cream, cake, pizza & ice-cream+beach, chocolate+cake, pizza+wine \\ \hline
\multicolumn{1}{|l|}{\textbf{People}} & woman, man, kid & woman+bag, man+boat, kid+dog \\ \hline
\end{tabular}
\label{tab:queries}
}
\end{center}
\end{table}

\begin{table}[h]
\noindent
\begin{minipage}[b]{0.48\linewidth}
\caption{Performance on InstaCities1M and WebVision. First column shows the mean P@5 for all the queries, second for the simple queries and third for complex queries.}
\centering
\resizebox{0.92\linewidth}{!}{
\begin{tabular}{|l|lll|lll|}
\hline
\rowcolor[HTML]{EFEFEF} 
\textbf{Text embedding} & \multicolumn{3}{c|}{\textbf{InstaCities1M}} & \multicolumn{3}{c|}{\textbf{WebVision}}\\ \hline
\rowcolor[HTML]{EFEFEF} 
\textbf{Queries} & \multicolumn{1}{c}{All} & \multicolumn{1}{c}{S} & \multicolumn{1}{c|}{C} & \multicolumn{1}{c}{All} & \multicolumn{1}{c}{S} & \multicolumn{1}{c|}{C} \\ \hline 
\textbf{LDA 200} & 0.40 & 0.73 & 0.07 & 0.11 & 0.18 & 0.03 \\ \cline{1-1}
\textbf{LDA 400} & 0.37 & 0.68 & 0.05 & 0.14 & 0.18 & 0.10 \\ \cline{1-1}
\textbf{Word2Vec mean} & 0.46 & 0.71 & \textbf{0.20} & 0.37 & 0.57 & 0.17 \\ \cline{1-1}
\textbf{Word2Vec tf-idf} & 0.41 & 0.63 & 0.18 & \textbf{0.41} & 0.58 & 0.23 \\ \cline{1-1}
\textbf{Doc2Vec} & 0.22 & 0.25 & 0.18 & 0.22 & 0.17 & \textbf{0.27} \\ \cline{1-1}
\textbf{GloVe} & 0.41 & 0.72 & 0.10 & 0.36 & \textbf{0.60} & 0.12 \\ \cline{1-1}
\textbf{GloVe tf-idf} & \textbf{0.47} & \textbf{0.82} & 0.12 & 0.39 & 0.57 & 0.22 \\ \cline{1-1}
\textbf{FastText tf-idf} & 0.31 & 0.50 & 0.12 & 0.37 & 0.60 & 0.13 \\ \hline 
\end{tabular}}
\vspace{0cm}
\label{tab:results}
\end{minipage}
\hspace{0.3cm}
\begin{minipage}[b]{0.47\linewidth}
\caption{Performance on transfer learning. First column shows the mean P@5 for all the queries, second for the simple queries and third for complex queries.}
\resizebox{\linewidth}{!}{
\centering
\begin{tabular}{|l|lll|lll|}
\hline
\rowcolor[HTML]{EFEFEF} 
\textbf{Text embedding} & \multicolumn{3}{|c|}{\textbf{\begin{tabular}[c]{@{}l@{}}Train: WebVision\\ Test: InstaCities\end{tabular}}} & \multicolumn{3}{c|}{\textbf{\begin{tabular}[c]{@{}l@{}}Train: InstaCities\\ Test: WebVision\end{tabular}}} \\ \hline
\rowcolor[HTML]{EFEFEF} 
\textbf{Queries} & \multicolumn{1}{c}{All} & \multicolumn{1}{c}{S} & \multicolumn{1}{c|}{C} & \multicolumn{1}{c}{All} & \multicolumn{1}{c}{S} & \multicolumn{1}{c|}{C} \\ \hline 
\textbf{LDA 200}  & 0.14 & 0.25 & 0.03 & 0.33 & 0.55 & 0.12 \\ \cline{1-1}
\textbf{LDA 400}  & 0.17 & 0.25 & 0.08 & 0.24 & 0.39 & 0.10 \\ \cline{1-1}
\textbf{Word2Vec mean}  & 0.41 & \textbf{0.63} & 0.18 & 0.33 & 0.52 & 0.15 \\ \cline{1-1}
\textbf{Word2Vec tf-idf}  & \textbf{0.42} & 0.57 & \textbf{0.27} & 0.32 & 0.50 & 0.13 \\ \cline{1-1}
\textbf{Doc2Vec}  & 0.27 & 0.40 & 0.15 & 0.24 & 0.33 & 0.15 \\ \cline{1-1}
\textbf{GloVe}  & 0.36 & 0.58 & 0.15 & 0.29 & 0.53 & 0.05 \\ \cline{1-1}
\textbf{GloVe tf-idf}  & 0.39 & 0.57 & 0.22 & \textbf{0.51} & \textbf{0.75} & \textbf{0.27} \\ \cline{1-1}
\textbf{FastText tf-idf}  & 0.39 & 0.57 & 0.22 & 0.18 & 0.33 & 0.03 \\ \hline
\end{tabular}}
\label{tab:results_transfer}
\end{minipage}

\hspace{0.3cm}
\centering
\caption{Performance on InstaCities1M using GloVe tf-idf introducing noise by changing the indicated \% of captions by random captions from the training set.\medskip}
\resizebox{0.38\linewidth}{!}{
\begin{tabular}{|l|lll|}
\hline
\rowcolor[HTML]{EFEFEF} 
\textbf{Experiment} & \multicolumn{3}{c|}{\cellcolor[HTML]{EFEFEF}\textbf{InstaCities1M}} \\ \hline
\rowcolor[HTML]{EFEFEF} 
\textbf{Queries} & \multicolumn{1}{c}{\cellcolor[HTML]{EFEFEF}All} & \multicolumn{1}{c}{\cellcolor[HTML]{EFEFEF}S} & \multicolumn{1}{c|}{\cellcolor[HTML]{EFEFEF}C} \\ \hline 
\textbf{Without introduced noise} & 0.47 & 0.82 & 0.12 \\ \cline{1-1}
\textbf{10\% introduced noise} & 0.25 & 0.43 & 0.07 \\ \cline{1-1}
\textbf{20\% introduced noise} & 0.18 & 0.32 & 0.05 \\ \cline{1-1}
\textbf{30\% introduced noise} & 0.15 & 0.25 & 0.05 \\ \cline{1-1} \hline 
\end{tabular}
\label{tab:results_noise}
}

\end{table}

\begin{figure}[h]
  \includegraphics[width=1\linewidth]{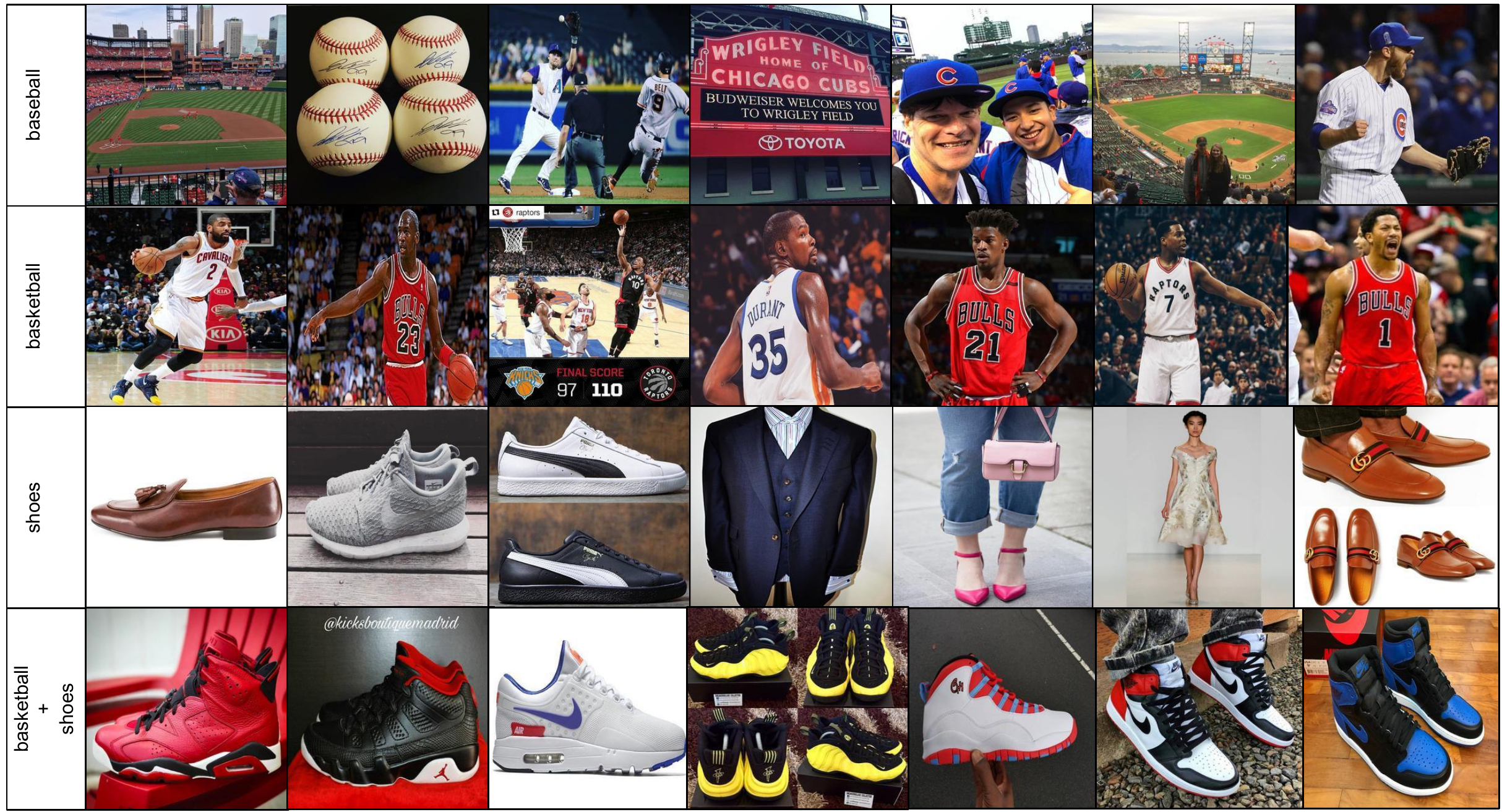}
   \caption{First retrieved images for complex queries with Word2Vec on InstaCites1M.}
   \label{fig:basketball}
\end{figure}

\begin{figure}[h]
  \includegraphics[width=1\linewidth]{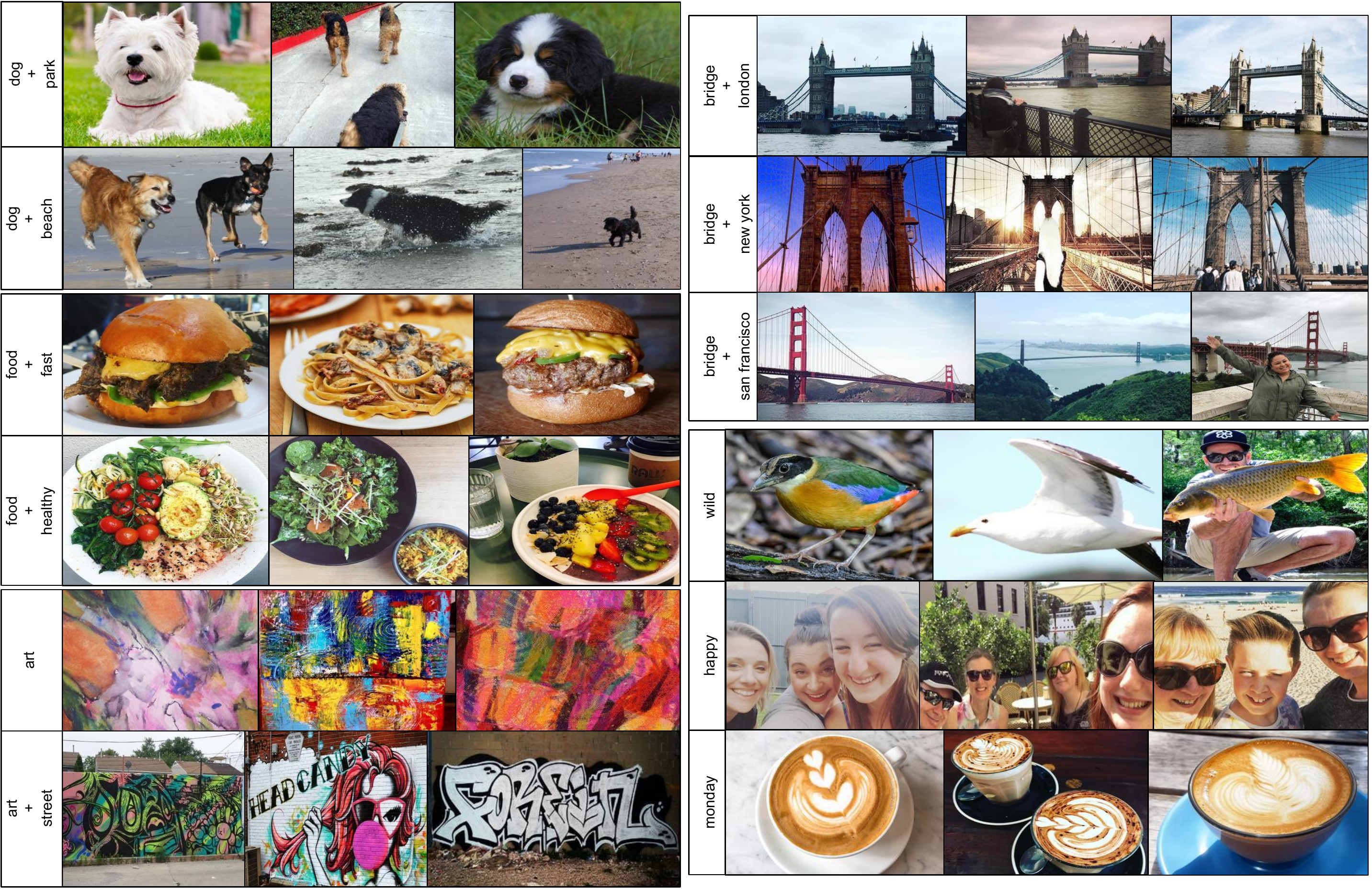}
   \caption{First retrieved images for complex queries (left), city related complex queries (top-right) and non-object queries (bottom-right) with Word2Vec on InstaCites1M.}
   \label{fig:queries_toguether}
\end{figure}

\begin{figure}[h]
  \includegraphics[width=1\linewidth]{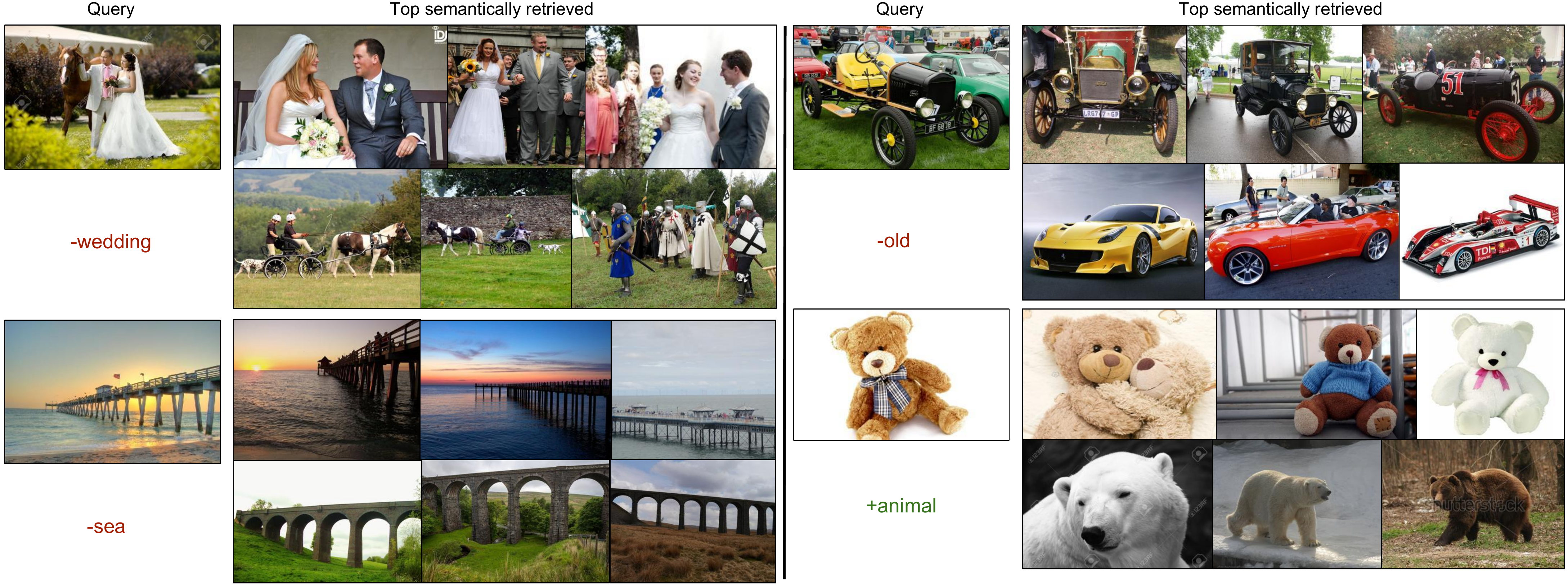}
   \caption{First retrieved images for multimodal queries (concepts are added or removed to bias the results) with Word2Vec on WebVision.}
   \label{fig:image_queries}
\end{figure}

\begin{figure}[h]
  \includegraphics[width=1\linewidth]{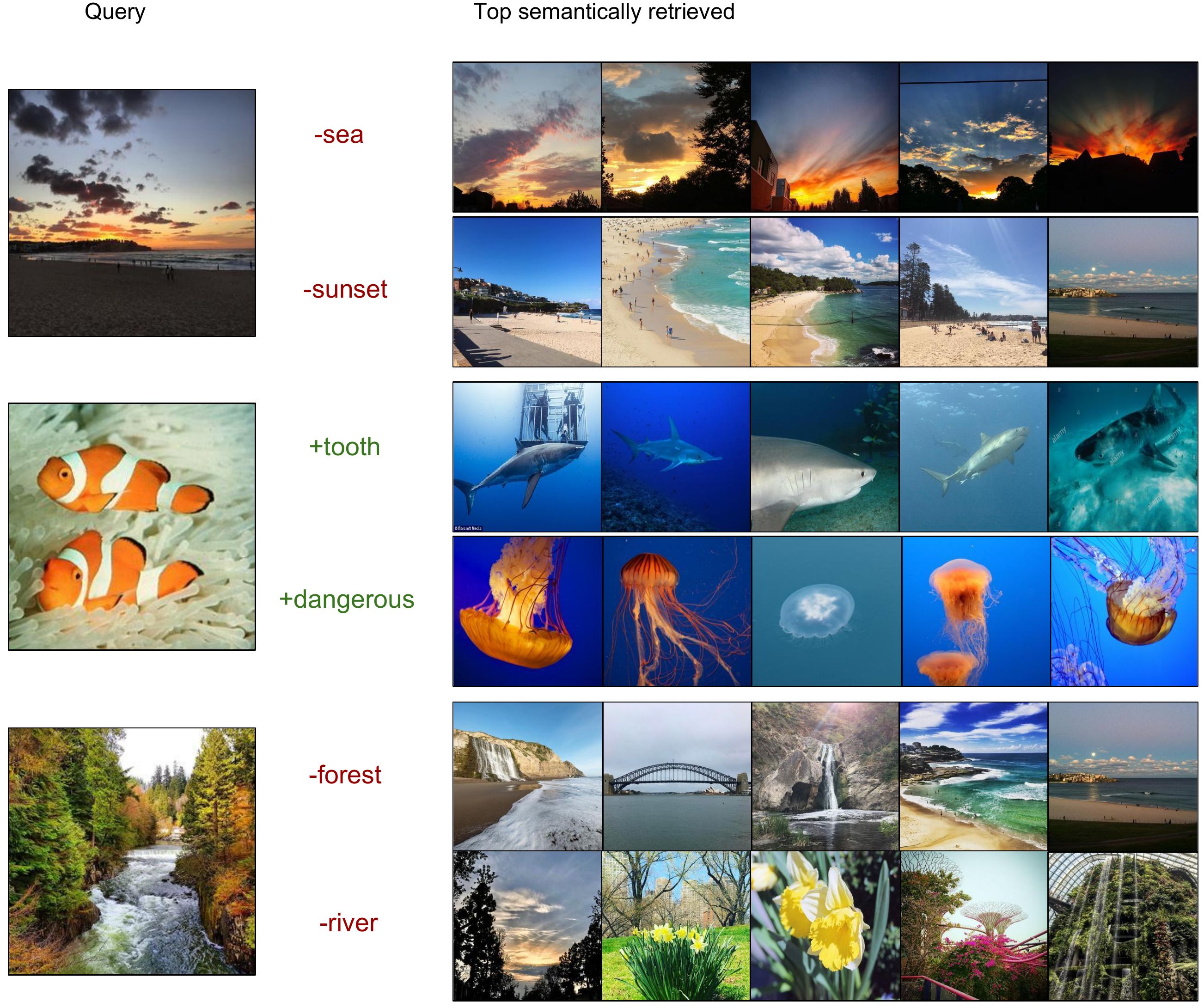}
   \caption{First retrieved images for multimodal complex queries with Word2Vec on WebVision.}
   \label{fig:image_queries_2}
\end{figure}

\subsection{Results and Conclusions}
Tables \ref{tab:results} and \ref{tab:results_transfer} show the mean Precision at 5 for InstaCities1M and WebVision datasets and transfer learning between those datasets. To compute transfer learning results, we train the model with one dataset and test with the other. 
Table \ref{tab:results_noise} shows the mean precision at 5 for InstaCities1M with introduced additional noise and of a model trained with Mean Square Error loss. The noise is introduced by changing the indicated \% of captions to random captions from the training set.
Figures \ref{fig:haircut}, \ref{fig:basketball} and \ref{fig:queries_toguether} show the first retrieved images for some complex textual queries. Figure \ref{fig:queries_toguether} also shows results for non-object queries, proving that our pipeline works beyond traditional instance-level retrieval.
Figures \ref{fig:image_queries} and \ref{fig:image_queries_2} show that retrieval also works with multimodal queries combining an image and text.

For complex queries, where we demand two concepts to appear in the retrieved images, we obtain good results for those queries where the concepts tend to appear together. For instance, we generally retrieve correct images for ``skyline + night'' and for ``bike + park'', but we do not retrieve images for ``dog + kid''. When failing with this complex queries, usually images where only one of the two querying concepts appears are retrieved. Figure \ref{fig:car_train} 
shows that in some cases images corresponding to semantic concepts between the two querying concepts are retrieved. That proves that the common embedding space that has been learnt has a semantic structure.  
The performance is generally better in InstaCities1M than in WebVision. The reason is that the queries are closer to the kind of images people tend to post in Instagram than to the ImageNet classes. However, the results on transfer learning show that WebVision is a better dataset to train than InstaCities1M. 
That's because WebVision has more images than InstaCities1M (2.4M training images vs 800k training images) and shows that the learned models are robust, general and scalable: Having more data, even if it's not specifically related with the target task, allows learning embedding models that perform better in that task.
Results show that all the tested text embeddings methods work quite well for simple queries. Though, LDA fails when is trained in WebVision. That is because LDA learns latent topics with semantic sense from the training data. Every WebVision image is associated to one of the 1,000 ImageNet classes, which influences a lot the topics learning. As a result, the embedding fails when the queries are not related to those classes. 
The top performing methods are GloVe when training with InstaCities1M and Word2Vec when training with WebVision, but the difference between their performance is small. 
FastText achieves a good performance on WebVision but a bad performance on InstaCities1M compared to the other methods. An explanation is that, while Social Media data contains more colloquial vocabulary, WebVision contains domain specific and diverse vocabulary, and since FastText learns representations for character ngrams, is more suitable to learn representations from corpus that are morphologically rich.
Doc2Vec does not work well in any database. That is because it is oriented to deal with larger texts than the ones we find accompanying images in Web and Social Media. 
For word embedding methods Word2Vec and GloVe, the results computing the text representation as the mean or as the \textit{tf-idf} weighted mean of the words embeddings are similar.\\
The overall conclusion of the performance comparison between text embeddings in this experiment is that word level text embeddings such as Word2Vec and GloVe perform better than document level text embeddings (LDA, Doc2Vec) and character ngrams level text embeddings (FastText). The reason is that captions associated to images in Social Media tend to be quite concise, so averaging the word-level embeddings of a caption still gives us an informative representation that allows us to take profit of the rich semantic space learnt by this kind of embeddings. The fact that this semantic space is quite sparse allows us to perform arithmetic between embeddings in it, and also to be able to learn from those representations averaged over caption's words.
The introduction of additional artificial noise deteriorates the results heavily. This indicates that, despite the proposed learning pipeline can learn powerful visual features from Web and Social Media data with its inherent noise, reducing it may lead to huge performance improvements. 

\subsection{Error Analysis}
Remarkable sources of errors are listed and explained in this section.

\subsubsection{Visual Features Confusion}
Errors due to the confusion between visually similar objects. For instance retrieving images of a quiche when querying ``pizza''. Those errors could be avoided using more data and a higher dimensional representations, since the problem is the lack of training data to learn visual features that generalize to unseen samples.

\subsubsection{Errors from the Dataset Statistics} 
\begin{figure}[h]
\begin{minipage}[c]{0.48\linewidth}\centering
  \includegraphics[width=1\linewidth]{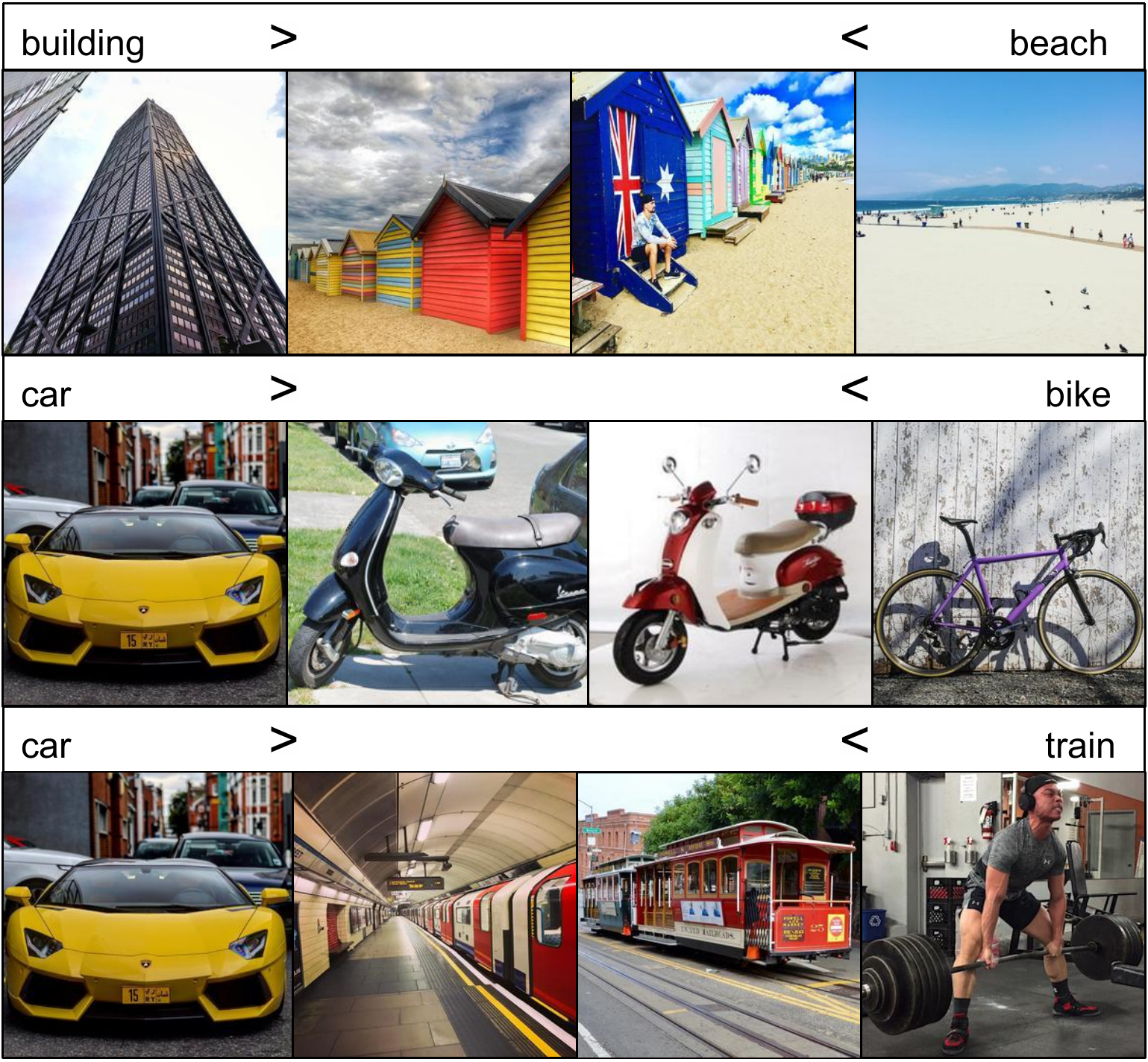}
   \caption{First retrieved images for simple (left and right columns) and complex weighted queries with Word2Vec on InstaCites1M.}
   \label{fig:car_train}
\end{minipage}
\hfill
\begin{minipage}[c]{0.48\linewidth}
\centering
   \includegraphics[width=1\linewidth]{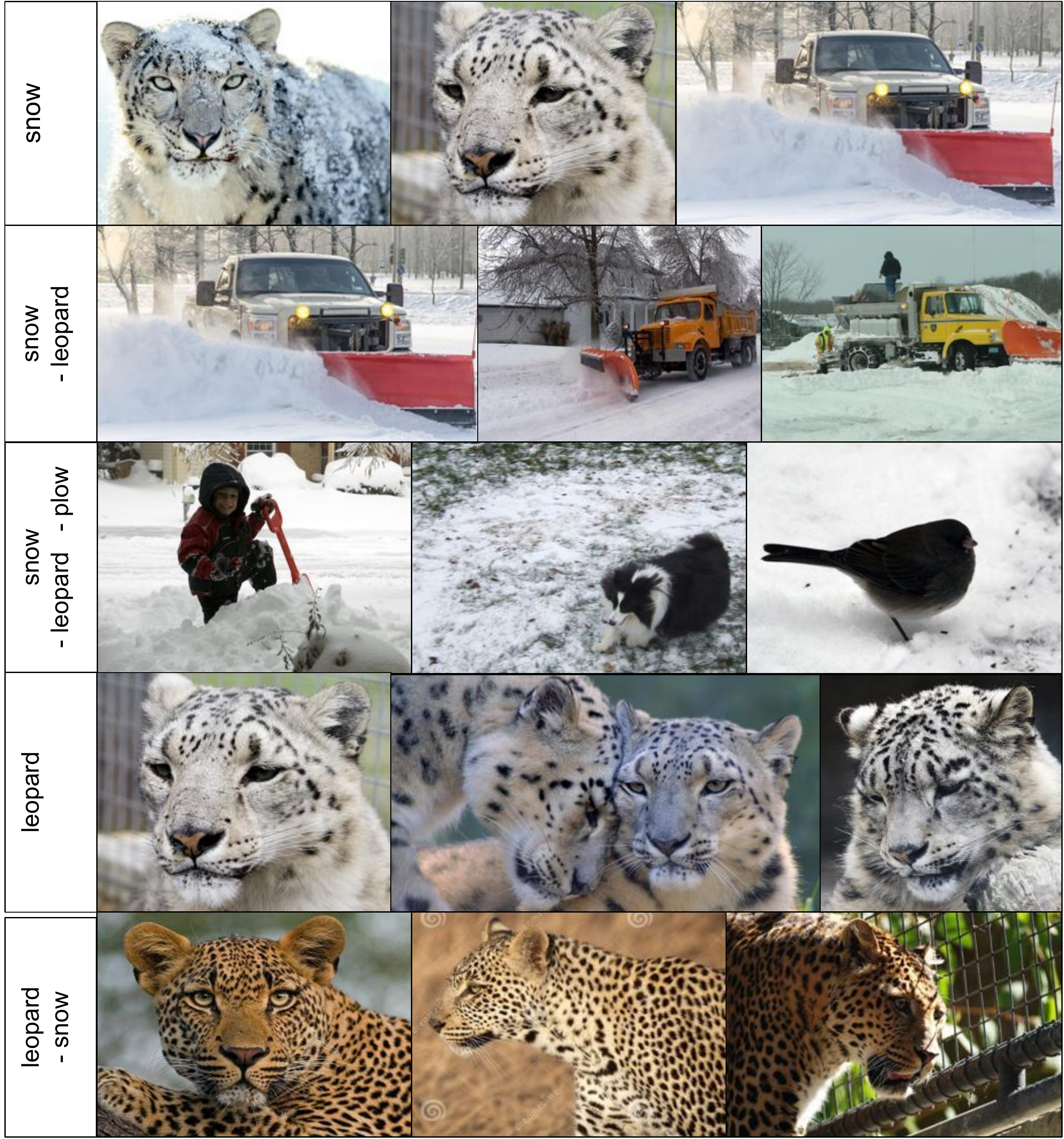}
   \caption{First retrieved images for text queries using Word2Vec on WebVision. Concepts are removed to bias the results.}
   \label{fig:snow_leopard}
\end{minipage}%
\end{figure}

An important source of errors is due to dataset statistics. As an example, the WebVision dataset contains a class which is ``snow leopard'' and it has many images of that concept. The word ``snow'' appears frequently in the images correlated descriptions, so the net learns to embed together the word ``snow'' and the visual features of a ``snow leopard''. There are many more images of ``snow leopard'' than of ``snow'', therefore, when we query ``snow'' we get snow leopard images. Figure \ref{fig:snow_leopard} shows this error and how we can use complex multimodal queries to bias the results.

\subsubsection{Words with Different Meanings or Uses} 
Words with different meanings or words that people use in different scenarios introduce unexpected behaviors. For instance when we query "woman + bag" in the InstaCities1M dataset we usually retrieve images of pink bags. The reason is that people tend to write "woman" in an image caption when pink stuff appears. 
Those are considered errors in our evaluation, but inferring which images people relate with certain words in Social Media can be a very interesting research.

\section{Retrieval in the MIRFlickr Dataset}
To compare the performance of our pipeline to other image retrieval by text systems we use the MIRFlickr dataset, which is typically used to train and evaluate image retrieval systems. The objective is to prove the quality of the multimodal embeddings learnt solely with Web data comparing them to supervised methods.\\

\begin{table}
\begin{tabular}{c c}
\begin{minipage}[t]{0.47\linewidth}

\centering
\caption{MAP on the image by text retrieval task on MIRFlickr as defined in \cite{Xu2017,Liu2017}.}
\resizebox{0.76\linewidth}{!}{
\begin{tabular}{|l|l|l|}
\hline
\rowcolor[HTML]{EFEFEF} \textbf{Method}          & \textbf{Train} & \textbf{map}   \\ \hline
\rowcolor[HTML]{FFFFFF} 
\textbf{LDA 200}  & InstaCites1M   & 0.736         \\ \hline
\textbf{LDA 400}  & WebVision      & 0.627          \\ \hline
\textbf{Word2Vec tf-idf} & InstaCites1M   & 0.720          \\ \hline
\textbf{Word2Vec tf-idf} & WebVision      & 0.738          \\ \hline
\textbf{GloVe tf-idf}    & InstaCites1M   & \textbf{0.756} \\ \hline
\textbf{GloVe tf-idf}    & WebVision      & 0.737          \\  \hline
\textbf{FastText tf-idf}    & InstaCities1M      & 0.677          \\  \hline
\textbf{FastText tf-idf}    & WebVision      & 0.734          \\ \hline \hline
\textbf{Word2Vec tf-idf}         & MIRFlickr   & 0.867      \\ \hline
\textbf{GloVe tf-idf}            & MIRFlickr   & \textbf{0.883}      \\ \hline
\textbf{DCH} \cite{Xu2017}             & MIRFlickr  & 0.813 \\ \hline
\textbf{LSRH} \cite{Li2016}            & MIRFlickr   & 0.768          \\ \hline
\textbf{CSDH} \cite{Liu2017}           & MIRFlickr   & 0.764          \\ \hline
\textbf{SePH} \cite{Lin2015}            & MIRFlickr   & 0.735          \\ \hline
\textbf{SCM} \cite{Zhang2014}             & MIRFlickr   & 0.631           \\ \hline
\textbf{CMFH} \cite{Ding2014}            & MIRFlickr  & 0.594           \\ \hline
\textbf{CRH} \cite{Zhen2012}             & MIRFlickr   & 0.581           \\ \hline
\textbf{KSH-CV} \cite{Zhou2014}          & MIRFlickr   & 0.571          \\ \hline
\end{tabular}
}
\label{tab:mirflickr}

\vspace{1.6cm}

\caption{MAP on the image by text retrieval task on MIRFlickr as defined in \cite{zhang2017}.}
\begin{center}
\vspace{-10pt}
\resizebox{0.66\linewidth}{!}{
\begin{tabular}{|l|l|l|}
\hline
\rowcolor[HTML]{EFEFEF} \textbf{Method}          & \textbf{Train} & \textbf{map}   \\ \hline
\rowcolor[HTML]{FFFFFF} 
\textbf{GloVe tf-idf}    & InstaCites1M   & \textbf{0.57} \\ \hline \hline
\textbf{GloVe tf-idf}            & MIRFlickr   & \textbf{0.73}      \\ \hline
\textbf{MML} \cite{zhang2017}             & MIRFlickr  & 0.63 \\ \hline
\textbf{InfR} \cite{zhang2017}            & MIRFlickr   & 0.60          \\ \hline
\textbf{SBOW} \cite{zhang2017}           & MIRFlickr   & 0.59          \\ \hline
\textbf{SLKL} \cite{zhang2017}             & MIRFlickr   & 0.55           \\ \hline
\textbf{MLKL} \cite{zhang2017}            & MIRFlickr  & 0.56           \\ \hline
\end{tabular}
}
\end{center}
\label{tab:mirflickr2}

\vspace{1.6cm}

\end{minipage}

\hspace{0.2cm}

& \begin{minipage}[t]{0.46\linewidth}

\caption{AP scores for 38 semantic concepts and MAP on MIRFlickr. Underlined numbers compare our method trained with InstaCities and other methods trained with the target dataset.}
\centering
\resizebox{\linewidth}{!}{
\begin{tabular}{|l||c|c|c||c|}
\hline
\rowcolor[HTML]{EFEFEF} 

\textbf{Method} 
& \begin{tabular}[c]{@{}c@{}}\textbf{GloVe}\\ \textbf{tf-idf}\end{tabular}   
& \begin{tabular}[c]{@{}c@{}}\textbf{MMSHL}\\ \cite{Wang2017}\end{tabular}  
& \begin{tabular}[c]{@{}c@{}}\textbf{SCM}\\ \cite{Zhang2014}\end{tabular}  
& \begin{tabular}[c]{@{}c@{}}\textbf{GloVe}\\ \textbf{tf-idf}\end{tabular} 
\\ \hline

\rowcolor[HTML]{EFEFEF} 
\textbf{Train} &  \multicolumn{3}{c||}{\textbf{MIRFlickr}}  & \textbf{InstaCities} \\ \hline
\rowcolor[HTML]{FFFFFF} 
\textbf{animals} & \textbf{0.775} & 0.382 & 0.353 & \underline{0.707} \\ \hline
\textbf{baby} & \textbf{0.337} & 0.126 & 0.127 & \underline{0.264} \\ \hline
\textbf{baby*} & \textbf{0.627} & 0.086 & 0.086 & \underline{0.492} \\ \hline
\textbf{bird} & \textbf{0.556} & 0.169 & 0.163 & \underline{0.483} \\ \hline
\textbf{bird*} & \textbf{0.603} & 0.178 & 0.163 & \underline{0.680} \\ \hline
\textbf{car} & \textbf{0.603} & 0.297 & 0.256 & \underline{0.450} \\ \hline
\textbf{car*} & \textbf{0.908} & 0.420 & 0.315 & \underline{0.858} \\ \hline
\textbf{female} & \textbf{0.693} & \underline{0.537} & 0.514 & 0.481 \\ \hline
\textbf{female*} & \textbf{0.770} & 0.494 & 0.466 & \underline{0.527} \\ \hline
\textbf{lake} & \textbf{0.403} & 0.194 & 0.182 & \underline{0.230} \\ \hline
\textbf{sea} & \textbf{0.720} & 0.469 & 0.498 & \underline{0.565} \\ \hline
\textbf{sea*} & \textbf{0.859} & 0.242 & 0.166 & \underline{0.731} \\ \hline
\textbf{tree} & \textbf{0.727} & \underline{0.423} & 0.339 & 0.398 \\ \hline
\textbf{tree*} & \textbf{0.894} & 0.423 & 0.339 & \underline{0.506} \\ \hline
\textbf{clouds} & \textbf{0.792} & \underline{0.739} & 0.698 & 0.613 \\ \hline
\textbf{clouds*} & \textbf{0.884} & 0.658 & 0.598 & \underline{0.710} \\ \hline
\textbf{dog} & \textbf{0.800} & 0.195 & 0.167 & \underline{0.760} \\ \hline
\textbf{dog*} & \textbf{0.901} & 0.238 & 0.228 & \underline{0.865} \\ \hline
\textbf{sky} & \textbf{0.900} & \underline{0.817} & 0.797 & 0.809 \\ \hline
\textbf{structures} & \textbf{0.850} & \underline{0.741} & 0.708 & 0.703 \\ \hline
\textbf{sunset} & \textbf{0.601} & \underline{0.596} & 0.563 & 0.590 \\ \hline
\textbf{transport} & \textbf{0.650} & \underline{0.394} & 0.368 & 0.287 \\ \hline
\textbf{water} & \textbf{0.759} & 0.545 & 0.508 & \underline{0.555} \\ \hline
\textbf{flower} & \textbf{0.715} & 0.433 & 0.386 & \underline{0.645} \\ \hline
\textbf{flower*} & \textbf{0.870} & 0.504 & 0.411 & \underline{0.818} \\ \hline
\textbf{food} & \textbf{0.712} & 0.419 & 0.355 & \underline{0.683} \\ \hline
\textbf{indoor} & \textbf{0.806} & \underline{0.677} & 0.659 & 0.304 \\ \hline
\textbf{plant \_life} & \textbf{0.846} & \underline{0.734} & 0.703 & 0.564 \\ \hline
\textbf{portrait} & \textbf{0.825} & \underline{0.616} & 0.524 & 0.474 \\ \hline
\textbf{portrait*} & \textbf{0.841} & \underline{0.613} & 0.520 & 0.483 \\ \hline
\textbf{river} & \textbf{0.436} & 0.163 & 0.156 & \underline{0.304} \\ \hline
\textbf{river*} & \textbf{0.497} & 0.134 & 0.142 & \underline{0.326} \\ \hline
\textbf{male} & \textbf{0.666} & \underline{0.475} & 0.469 & 0.330 \\ \hline
\textbf{male*} & \textbf{0.743} & \underline{0.376} & 0.341 & 0.338 \\ \hline
\textbf{night} & \textbf{0.589} & \underline{0.564} & 0.538 & 0.542 \\ \hline
\textbf{night*} & \textbf{0.804} & 0.414 & 0.420 & \underline{0.720} \\ \hline
\textbf{people} & \textbf{0.910} & \underline{0.738} & 0.715 & 0.640 \\ \hline
\textbf{people*} & \textbf{0.945} &  \underline{0.677} & 0.648 & 0.658 \\ \hline
\rowcolor[HTML]{EFEFEF} 
\textbf{MAP} & \textbf{0.738} & 0.451 & 0.415 & \underline{0.555} \\ \hline
\end{tabular}}
\label{tab:query_classes}
\end{minipage}

\end{tabular}
\end{table}

\subsection{Experiment Setup}
We consider three different experiments: 
1) Using as queries the tags accompanying the query images and computing the MAP of all the queries. Here a retrieved image is considered correct if it shares at least one tag with the query image. For this experiment, the splits used are 5\% queries set and 95\% training and retrieval set, as defined in \cite{Xu2017,Liu2017}.
2) Using as queries the class names. Here a retrieved image is considered correct if it is tagged with the query concept. For this experiment, the splits used are 50\% training and 50\% retrieval set, as defined in \cite{Wang2017}.
3) Same as experiment~1 but using the MIRFlickr train-test split proposed in Zhang et al. \cite{zhang2017}.\\

\subsection{Results and Conclusions}
Tables \ref{tab:mirflickr} and \ref{tab:mirflickr2} show the results for the experiments 1 and 3 respectively. We see that our pipeline trained with Web and Social Media data in a multimodal self-supervised fashion achieves competitive results. When trained with the target dataset, our pipeline outperforms the other methods. Table \ref{tab:query_classes} shows results for the experiment 2. Our pipeline with the GloVe \textit{tf-idf} text embedding trained with InstaCites1M outperforms state of the art methods in most of the classes and in MAP. If we train with the target dataset, results are improved significantly. 
Notice that despite being applied here to the classes and tags existing in MIRFlickr, our pipeline is generic and has learnt to produce joint image and text embeddings for many more semantic concepts, as seen in the qualitative examples.\\

\section{Comparing the Image and Text Embeddings}
In this section we analyze the semantic quality of the learnt joint embedding spaces showing how the CNN has learnt to embed images in them.

\begin{figure*}[h]
	\centering
  \includegraphics[width=\linewidth]{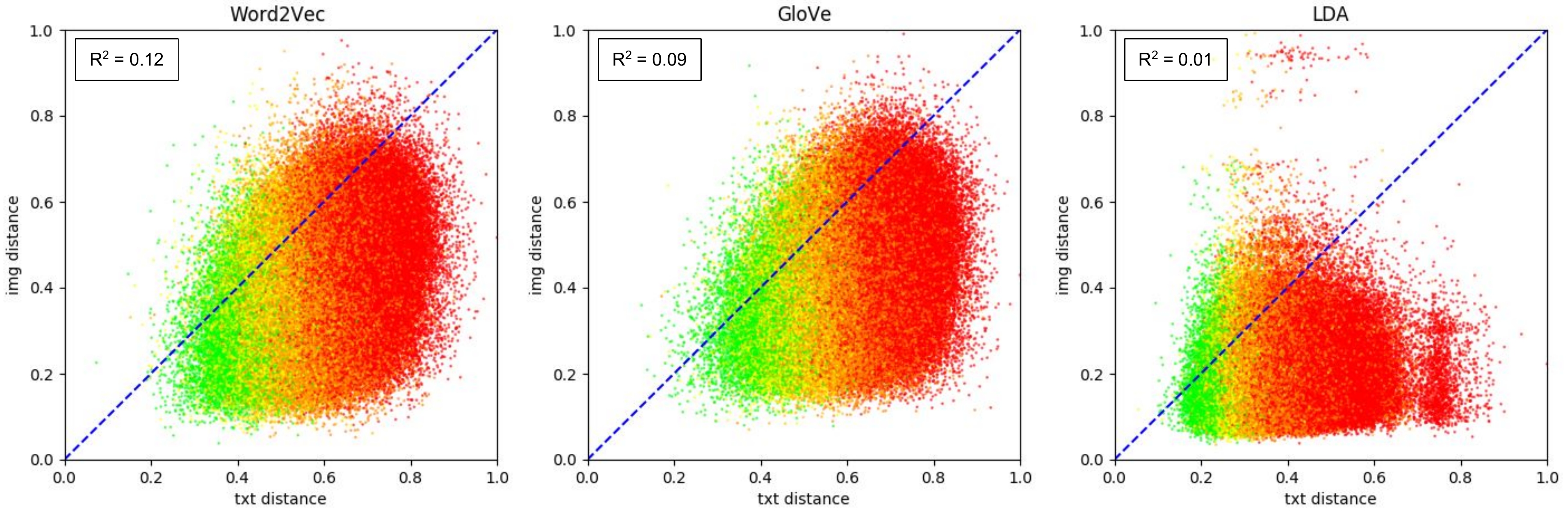}
   \caption{Text embeddings distance (X) vs the images embedding distance (Y) of different random image pairs for LDA, Word2Vec and GloVe embeddings trained with InstaCities1M. Distances have been normalized between [0,1]. Points are red if the  pair does not share any tag, orange if it shares 1, light orange if it shares 2, yellow if it shares 3 and green if it shares more. R\textsuperscript{2} is the coefficient of determination of images and texts distances.}
   \label{fig:embeddings}
\end{figure*}

\subsection{Experiment Setup}
To evaluate how the CNN has learnt to map images to the text embedding space and the semantic quality of that space, we perform the following experiment: We build random image pairs from the MIRFlickr dataset and we compute the cosine similarity between 
both their image and their text embeddings.
In Figure \ref{fig:embeddings} we plot the images embeddings distance vs the text embedding distance of 20,000 random image pairs. 
If the CNN has learnt correctly to map images to the text embedding space, the distances between the embeddings of the images and the texts of a pair should be similar, and points in the plot should fall around the identity line $y=x$. Also, if the learnt space has a semantic structure, both the distance between images embeddings and the distance between texts embeddings should be smaller for those pairs sharing more tags: The plot points' color reflects the number of common tags of the image pair, so pairs sharing more tags should be closer to the axis origin.

As an example, take a dog image with the tag "dog", a cat image with the tag "cat" and one of a scarab with the tag "scarab". If the text embedding has been learnt correctly, the distance between the projections of dog and scarab tags in the text embedding space should be bigger than the one between dog and cat tags, but smaller than the one between other pairs not related at all.
If the CNN has correctly learnt to embed the images of those animals in the text embedding space, the distance between the dog and the cat image embeddings should be similar than the one between their tags embeddings (and the same for any pair). So the point given by the pair should fall in the identity line. 
Furthermore, that distance should be nearer to the coordinates origin than the point given by the dog and scarab pair, which should also fall in the identity line and nearer to the coordinates origin that another pair that has no relation at all.\\

\subsection{Results and Conclusions}
The plots in Figure \ref{fig:embeddings} for both the Word2Vec and the GloVe embeddings show a similar shape. The resulting blob is elongated along the $y=x$ direction, which proves that both image and text embeddings tend to provide similar distances for an image pair. The blob is thinner and closer to the identity line when the distances are smaller (so when the image pairs are related), which means that the embeddings can provide a valid distance for semantic concepts that are close enough (dog, cat), but fails inferring distances between weak related concepts (car, skateboard). 
The colors of the points in the plots show that the space learnt has a semantic structure. Points corresponding to pairs having more tags in common are closer to the coordinates origin and have smaller distances between the image and the text embedding. From the colors it can also be deducted that the CNN is good inferring distances for related images pairs: there are just a few images having more than 3 tags in common with image embedding distance bigger than 0.6, while there are many images with bigger distances that do not have tags in common. However, the visual embedding sometimes fails and infers small distances for image pairs that are not related, as those images pairs having no tags in common and an image embedding distance below 0.2.

The plot of the LDA embedding shows that the learnt joint embedding is not so good in terms of the CNN images mapping to the text embedding space nor in terms of the space semantic structure. The blob does not follow the identity line direction that much which means that the CNN and the LDA are not inferring similar distances for images and texts of pairs. The points colors show that the CNN is inferring smaller distances for more similar image pairs only when the pairs are very related.

The coefficient of determination R\textsuperscript{2} shown at each graph measures the proportion of the variance in a dependent variable that is predicted by linear regression and a predictor variable. In this case, it can be interpreted as a measure of how much image distances can be predicted from text distances and, therefore, of how well the visual embedding has learnt to map images to the joint image-text space. It ratifies our plots' visual inspection proving that visual embeddings trained with Word2Vec and GloVe representations have learnt a much more accurate mapping than LDA, and shows that Word2Vec is better in terms of that mapping. 

\section{Visualizing CNN activation maps}
\begin{figure}[h]
\begin{center}
  \includegraphics[width=0.6\linewidth]{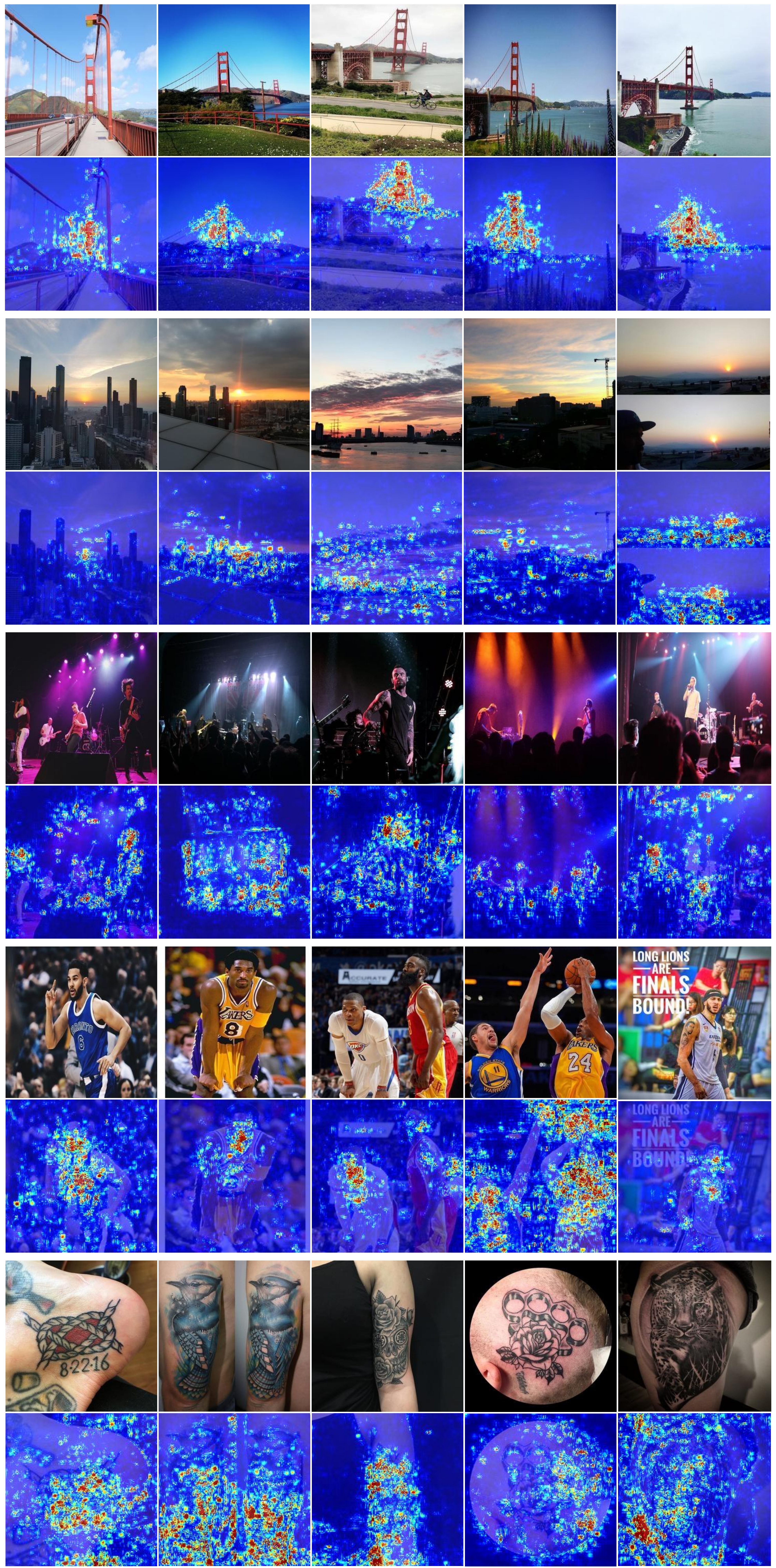}
     \caption{Top-5 activations for five units in \text{pool5} layer of GoogleNet model trained from scratch with InstaCities1M using GloVe tf-idf as self-supervision and their activation maps.}
  \label{fig:activation_maps}
  \end{center}
\end{figure}

We have proved that, using only Social Media data, state of the art CNNs can be trained in a self-supervised way to learn powerful visual features, capable to discriminate among a huge variety of scenes: from objects to outdoor scenes, abstract concepts or specific buildings.
In this experiment we visualize the images from the InstaCities1M retrieval set that generated the highest activations in some CNN units, using the GoogleNet trained from scratch with InstaCites1M and GloVe tf-idf text embedding as self-supervision. We also show the regions of the images that activated most the selected units. To generate those activations maps we used \textit{deconvnet}, proposed by Zeiler et al. \cite{Zeiler2014} and the Caffe implementation presented in \cite{Yosinski2}. Figure \ref{fig:activation_maps} shows the results of a selection of neurons in the \textit{pool5} layer of our model. We can notice that network units are selective to specific buildings, such as Golden Gate Bridge, objects such as guitars, drums or lights to identify concert scenes, or even basketball t-shirts.

\section{Visualizing the Learned Semantic Space with t-SNE}
In this section we use the t-SNE dimensionality reduction method to reduce the dimensionality of the joint embedding space to 2 dimensions and we show images in that space to visualize its semantic structure. 

\subsection{Dimensionality Reduction with t-SNE}
Inspired by A. Karpathy's work\footnote{\url{https://cs.stanford.edu/people/karpathy/cnnembed/}}, who uses t-SNE to visualize CNN layer features, we use t-SNE\footnote{\url{https://github.com/lvdmaaten/bhtsne/}} \cite{VanDerMaaten2014} to visualize the learnt joint visual and textual embedding. t-SNE is a non-linear dimensionality reduction method, which we use on our 400 dimensional embeddings to produce 2 dimensional embeddings. 
For each one of the given 400 dimensional visual or textual embeddings, t-SNE computes a 2 dimensional embedding arranging elements that have similar representations nearby, providing a way to visualize the learnt joint image-text space and analyze qualitatively its semantic structure.
\subsection{Visualizing both image and text embeddings}
As we have learnt a joint image and text embedding space, we can apply t-SNE to both modalities of embeddings at once. We apply t-SNE to a set formed by the visual embeddings of the images in test set of InstaCities1M and the text embeddings of the selected querying terms (Table \ref{tab:queries}). In this experiment, we use the Word2Vec model trained on InstaCities1M dataset.

\subsection{Showing Images at the Embedding Locations}
First, we set a canvas with predefined dimensions (2000x2000 pixels). Then we normalize the 2 dimensional embeddings given by t-SNE to fit in the canvas size. Finally, we visualize images at their embedding locations, setting their top-left corner at their embedding location and resizing them to 50x50 pixels. For text embeddings, we use an image containing its words as their representations in the canvas. 
To get an interpretable visualization avoiding images overlaps, if two images share any pixel in the output figure we omit one of them (prioritizing word images). Therefore, images surrounding word images are not necessary top retrieval results for that word, but they are the nearest images of the ones being represented in the figure.
\subsection{Semantic Space Inspection}
\begin{figure}[h]
  \includegraphics[width=\linewidth]{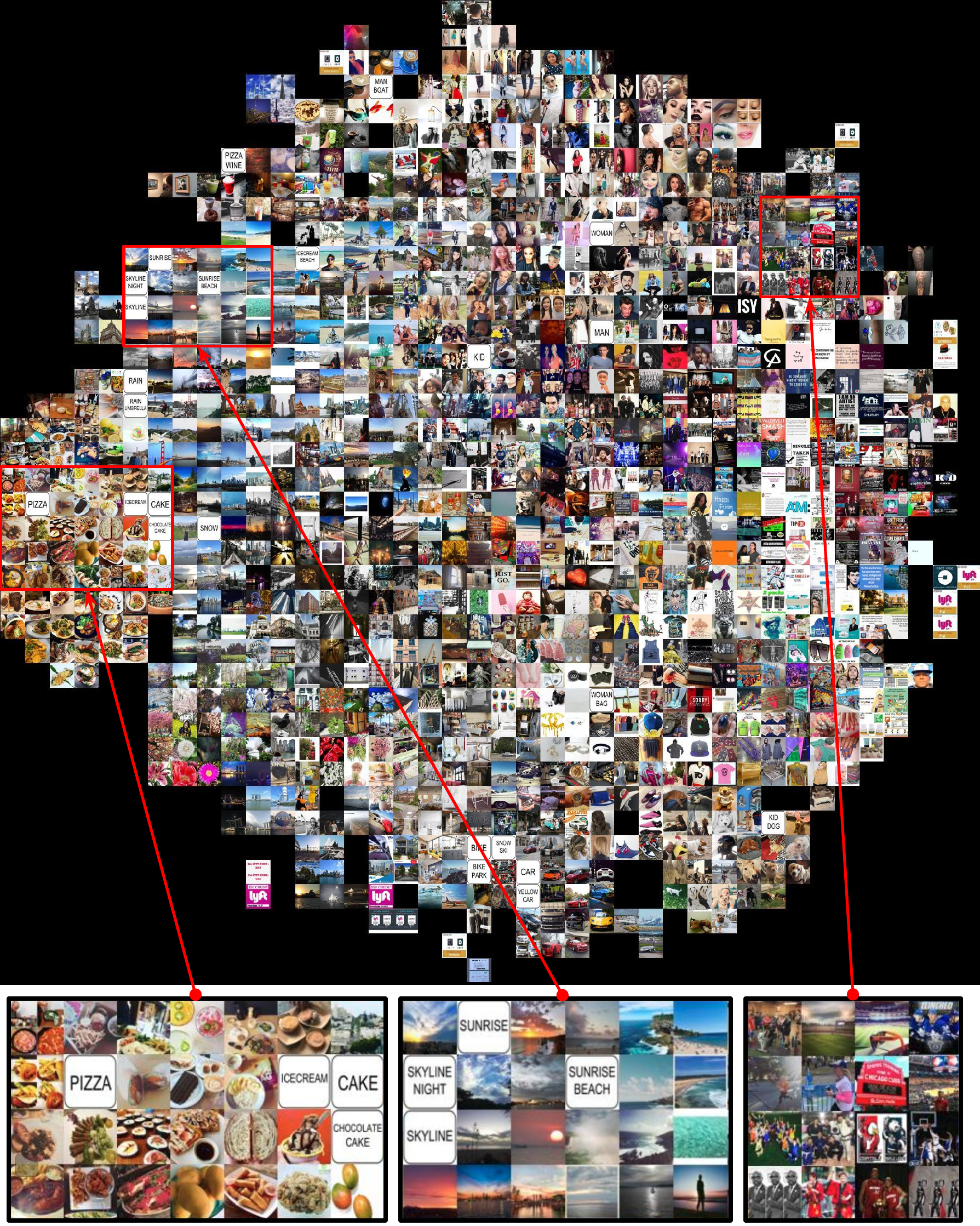}
     \caption{Visualization (2000x2000 px) of the joint embedding with Word2Vec on InstaCities1M dataset.}
  \label{fig:tsne}
\end{figure}

The joint embeddings 2 dimensional visualization in \ref{fig:tsne} shows the semantic structure of the learnt space.
It shows semantic clusters that the joint embedding has learnt in a self-supervised way from the data distribution, that correspond to different kind of images people tend to post on Instagram. For instance, the figure shows a cluster for food images, a cluster for sport images, a cluster for sunrise images, or a cluster for animal images. It also shows that images of people are very numerous, and that the joint embedding groups them correctly. It can also be appreciated how images we might consider noise, such as images with logos or text, are clustered together. The majority of those images are far from the semantic clusters, isolated and near the figure edges. That is because the joint embedding hasn't been able to find semantic relations between these images and the rest, so it assigns to them embeddings that have not relation with the others. When computing t-SNE, as the objective is to place similar images nearby, this images without semantic relations are set far from the others. Therefore, we can conclude that the pipeline is quite robust to Social Media noise. 
More t-SNE visualizations of the learnt joint embeddings are avalaible in \url{https://gombru.github.io/2018/08/01/learning_from_web_data}.

\section{Conclusions}
In this work we learn a joint visual and textual embedding using Web and Social Media data and we benchmark state of the art text embeddings in the image retrieval by text task, concluding that GloVe and Word2Vec are the best ones for this data, having a similar performance and competitive performances over supervised  methods in the image retrieval by text task. 
We show that our models go beyond instance-level image retrieval to semantic retrieval and that can handle multiple concepts queries and also multimodal queries, composed by a visual query and a text modifier to bias the results.
We clearly outperform state of the art in the MIRFlick dataset when training in the target data. 
The code used in this work is available on \url{https://github.com/gombru/LearnFromWebData}.

\section*{Acknowledgments}
This work was supported by the Doctorats Industrials program from the Generalitat de Catalunya, the Spanish project TIN2017-89779-P, the H2020 Marie Skłodowska-Curie actions of the European Union, grant agreement No 712949 (TECNIOspring PLUS), and the Agency for Business Competitiveness of the Government of Catalonia (ACCIO).

\clearpage
\clearpage




\bibliographystyle{elsarticle-num}

\bibliography{mybib}
\clearpage

\end{document}